Short Paper*

# GULP: Solar-Powered Smart Garbage Segregation Bins with SMS Notification and Machine Learning Image Processing

Jerome B. Sigongan
College of Computer Studies, Northern Bukidnon State College, Philippines
20191125@nbsc.edu.ph

Hamer P. Sinodlay
College of Computer Studies, Northern Bukidnon State College, Philippines
20191178@nbsc.edu.ph

Shahida Xerxy P. Cuizon
College of Computer Studies, Northern Bukidnon State College, Philippines
20191011@nbsc.edu.ph

Joanna S. Redondo
College of Computer Studies, Northern Bukidnon State College, Philippines
20191440@nbsc.edu.ph

Maricel G. Macapulay
College of Computer Studies, Northern Bukidnon State College, Philippines
20191515@nbsc.edu.ph

Charlene O. Bulahan-Undag
College of Computer Studies, Northern Bukidnon State College, Philippines
cobulahan@nbsc.edu.ph

Kenn Migan Vincent C. Gumonan
College of Computer Studies, Northern Bukidnon State College, Philippines
kmvcgumonan@nbsc.edu.ph
(corresponding author)

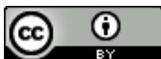

**Abstract**

*Purpose* – This study intends to build a smart bin that segregates solid waste into its respective bins. To make the waste management process more interesting for the end-users; to notify the utility staff when the smart bin needs to be unloaded; to encourage an environment-friendly smart bin by utilizing renewable solar energy source.

*Method* – The researchers employed an Agile Development approach because it enables teams to manage their workloads successfully and create the highest-quality product while staying within their allocated budget. The six fundamental phases are planning, design, development, test, release, and feedback.

*Results* – The Overall quality testing result that was provided through the ISO/IEC 25010 evaluation which concludes a positive outcome. The overall average was 4.55, which is verbally interpreted as excellent.

*Conclusion* – The researchers were able to develop an innovative smart bin that automates the solid waste disposal in the target institution and also monitors the garbage bins to prevent overflow through its SMS notification feature. Additionally, the application can also independently run with its solar energy source. Users were able to enjoy the whole process of waste disposal through its interesting mechanisms.

*Recommendations* – Based on the findings, a compressor is recommended to compress the trash when the trash level reaches its maximum point to create more rooms for more garbage. An algorithm to determine multiple garbage at a time is also recommended. Adding a solar tracker coupled with solar panel will help produce more renewable energy for the smart bin.

*Research Implications* – The institution continues to operate on the basis of traditional waste management along with the increasing student population, and an increasing effort in monitoring proper waste disposal. The traditional method of manually separating waste in a garbage bin is time-consuming and costly. Having an efficient and sustainable waste management application will aid the regulation of waste disposal and ease some environmental pressures.

## INTRODUCTION

Solid waste management is a growing challenge that has direct impact. Annually, the world produces 2.01 billion metric tons of municipal solid waste, with at least 33% of that not properly handled for the environment. The amount of waste generated by each individual every day across the world averages 0.74 kilograms but varies significantly, going from 0.11 to 4.54 kilograms. By 2050, global consumption is anticipated to amount to 3.40 billion metric tons (Kaza et al., 2018). Waste segregation is the practice of sorting waste products from other trash in order to reuse, recycle, and reduce waste (Kihila et al., 2021). Although people have always generated waste, waste management has become more difficult as industry and technology have advanced (Abdel-Shafy & Mansour, 2018). Most developing countries lacks in the implementation of waste segregation (Hondo et al., 2020). As reported by the Department of Environment and Natural Resources (DENR), garbage management continues to be a major issue in the Philippines, due primarily to inconsistent waste segregation at the local community level. Although barangay-level segregation is required under Republic Act 9003 (commonly known as the Ecological Solid Waste Management Act of 2000), certain local government units (LGUs) have not yet rigorously enforced it. Meanwhile, there is still a lack of recognition among households about the value of waste segregation, resulting in no changes in disposal habits (Philippine News Agency, 2022).

The waste management method is riddled with flaws. Waste is taken from garbage cans at every colony; however, failing to collect them on a specific day rather than based on their status has the disadvantage of recurrently overflowing bins, which makes litter bins ideal for microbial growth, animal feeding, and insect mating (Gangwani et al., 2019). In confronting environmental challenges, academic institutions play an important role as centers of learning. They aid in the growth of environmental knowledge, abilities, and values, which will effectively change the way that staff members and students think about careless trash disposal at work and at home. Through conservation and sustainability, academic institutions have made a substantial contribution to creating and sustaining the quality of communal and human life on campus (Madrigal & Oracion, 2011).

The study aims to develop a smart waste segregator system powered by solar panel as a sustainable energy source. The garbage disposal system will be able to perform real-time object recognition and categorization of the garbage if it is biodegradable, non-biodegradable, or recyclable. The study also aims to create a module that will notify the end-users if the garbage bins are full and create an interactive way for the users to throw their garbage.

## LITERATURE REVIEW

The authors employed linked published resources that will serve as a guide in reaching the target objectives while referencing other similar systems and making adjustments as necessary in order to better comprehend the research. A university in Pune, India, developed a smart dustbin that automatically separates dry and wet waste. The smart waste-bin design is based on both software and hardware implementation. Arduinos, sensors, servo motors, and batteries are among the hardware components used. The smart waste-bin has two main steps: motion detection and dry-wet waste sorting. The project was successful in meeting the lid opening scenario and segregating dry and wet waste. To avoid human contact or interference, motion detection is used. The researchers suggested using solar panels to make the Smart Waste Bin more energy efficient. Similarly, the proposed bin currently has two compartments in a single container; however, the bin design is recommended for redesign so that different types of waste can be separated into different containers. Furthermore, suggesting AI features for the system to make waste management even easier (Kulkarni et al., 2020).

A study was carried out by Pimpri Chinchwad College of Engineering and Research at Ravet, Pune, India, on an autonomous waste segregator and monitoring system. An ATmega328 serves as the main controller for the segregation section. An inductive proximity sensor, a capacitive proximity sensor, and a moisture sensor were integrated in order to determine the sort of trash. By using an automatic garbage segregator, the waste is divided into categories such as plastic, dry, moist, or metallic, and the monitoring section oversees the waste collection procedure. With the hardware they used, they were able to create a simple, low-cost system. The use of object detection through image processing was suggested by the researchers to progress the research. This approach may recognize the incoming items by using a database of the objects that has already been fed (Mapari et al., 2020). In another study, a Smart Trash Segregator Dustbin Monitoring System was sought after by Alva's Institute of Engineering and Technology in Mijar, Karnataka, India. At the source level, the project may separate wet and dry garbage and then alert the Municipal Corporation. The process of rubbish collecting was observed using an IR sensor. A signal is transmitted to an Arduino UNO and information is communicated to the Municipal Corporation through a Wemos board when the sensor detects that rubbish has reached the sensors' level. They underlined in this study the local government's worries regarding the labor-intensive process of destination-level waste segregation. The experts advised sorting rubbish at the source in order for the Municipal Corporation to appropriately dispose of and recycle it (Sharma & Singh, 2018). A similar study from Can Tho University in Vietnam aimed at transforming their institution's normal trash bin into a smarter one by using computer vision technology. The trash bin can automatically classify garbage with the help of sensor and actuator devices. A camera mounted on the trash bin captures images of trash, which are then analyzed and decided upon by the Raspberry Pi central processing unit. Additionally, the model needs an Internet connection to update the bin status using the Firebase real-time database for additional control via a mobile application created using React-Native (Lam et al., 2021). The Sri Lanka Institute of Information Technology has devised a method for enabling society to use Internet of Things (IOT) technologies to automate and optimize waste management procedures. The suggested solution uses an Alert Notification Sub System to alert the user and other authorities when the waste bins are full or if there is an unusual condition inside the bins, such as a high temperature or high humidity. Moreover, the authors used sound and light indicators to determine bin status with dashboard (Kumari et al., 2019).

The design and development of a smart green environment garbage monitoring system that measures garbage levels in real time and notifies the municipality when the bin is full based on the types of garbage was presented in a project titled "Smart Bin: Internet-of-Things Garbage Monitoring System" from Universiti Malaysia Perlis in Arau, Perlis, Malaysia. The suggested system is made up of ultrasonic sensors that gauge the quantity of rubbish, an ARM microcontroller that manages system operations, and everything will be connected to ThingSpeak to store data for later usage and analysis, such as forecasting the peak level of garbage bin fullness. However, the researchers pointed out that if the network has limited coverage or availability, the WiFi module will not function, causing the entire system to malfunction. As a result, a development of communication hardware is required in the future. To prevent accidentally putting waste in the wrong bin, they also suggested creating a system that uses image recognition to recognize various sorts of trash. Indirectly, this will lessen traffic congestion and save money and labor (Mustafa & Ku Azir, 2017). A study on the IOT Based Smart Garbage Monitoring System Using Raspberry Pi with GPS Link was carried out by Anna University in Tamil Nadu, India. The device is an Internet of Things-based smart waste monitoring system with GPS connectivity, real-time monitoring, and alarm features. The Raspberry-Pi module, GSM module, and GPS antenna were all used by the researchers. Furthermore, they used a bin compression technique that compresses the bin when it is overloaded and notifies the user that it has been compressed (Tejaswini et al., 2019). Students from LPU-Laguna designed and implemented an efficient system to help properly segregate wastes in their study "Design and Implementation of Automated Trash Bin with Smart Compression." All of the parts needed to produce the desired output are linked together with two Arduino Uno. To choose the type of garbage to be disposed of, push buttons are employed. To show whether the input was valid, LEDs were employed. The trash cans that are fixed to a circular metal plate are rotated by the stepper motor. To identify whether paper or plastic bins were necessary for compression, an ultrasonic sensor was utilized. Proponents strongly advised using image processing in the system for further development. This will result in a system that is not reliant on human knowledge (Paul et al., 2020).

A disguised automated garbage can with a solid waste shredder was suggested in a study by the Camarines Sur Polytechnic Colleges. The trash cans are made to be hidden in order to conserve space, slow the decomposition of trash, and lessen trash odor. In order to minimize direct garbage interaction, the design is totally automated. Waste can be gathered and separated using a gripper, servo motors, ultrasonic, capacitive, and photoelectric sensors. To save energy, the separated waste is kept in a storage bin before shredding. Waste is then transported to the appropriate transport bins for collection after shredding. The GSM module can send an SMS to the user notifying them of the transport bins' capacity thanks to the ultrasonic sensors. The design was found to be workable and may be implemented utilizing inexpensive and efficient electrical components such capacitive and photoelectric sensors, GSM modules, ultrasonic sensors, AC and servo motors, and Arduino Uno and Mega microcontrollers. However, the prototype took significantly too long to collect, categorize, and segregate waste in compared to the predicted reaction times, the researchers discovered after utilizing efficient yet affordable electronic gadgets The researchers suggested that in order to obtain a shorter response time, the system be improved with more advanced automation capabilities, such as an image recognition algorithm, for quicker and more accurate object detection and classification (Villamer et al., 2022). Proponents from Noida International University in Greater Noida, India, built a project to prevent trash cans from being overfilled. The project entailed giving trash cans the ability to detect when they needed to be cleaned. The smart trash can management system is based on ESP8266 Wi-Fi module and has ultrasonic sensors on each trash can to display the garbage's current condition on an LCD screen and a mobile device. When used, the smart bin provides an incentive by polishing the shoe using the PIR sensor, APR module, and IR sensor (Chaturvedi et al., 2021).

The study "The Design and Implementation of Smart Trash Bin" provided a cost-effective design of an intelligent waste container for small-scale applications. This system employed an Arduino Nano board, an ultrasonic sensor, and a GSM module to monitor the container's level of fullness and deliver SMS alerts. A lithium battery power bank and a solar cell panel work together to power the system. The power bank can be used by the system to charge external portable devices. Additionally, the system records usage and fullness events for the bin while it is in use on a memory card, which is also used to play an audio message through a speaker (Fadel, 2017). At the Southern Philippines Agri-Business and Marine and Aquatic School of Technology in Digos City, the Philippines, a similar project was also created. The main objectives of the project were to develop a monitoring system that uses an ultrasonic sensor to show the level of garbage in the trash cans in real-time, a GSM module to alert the administrator when the level of garbage in the trash can reaches the maximum level, and a garbage can that opens on its own using a servomotor and laser sensor when it detects nearby objects. The system's back end was a Text File, and the project team used Visual Studio 2012 as the front end.

Asp.Net was used to construct the user interface. The project team used these programs to produce a desktop application that graphs the amount of trash present in the trash cans and presents that information (Padal Jr et al., 2019). In order to help consumers, make the best use of the trash can, proponents of the study "Smart Waste Management System Using Internet of Things and RFID Technology" developed a smart rubbish. With the aid of a micro-controller, sensors, servomotors, a DC motor, RFID, and Wi-Fi technologies, this amazing garbage disposal system was created. The project's objective was to inform and honor the various users who daily dump trash. The program comprises a Wi-Fi system that can transmit user data to an IOT-enabled application as well as an RFID learning module that can monitor user identification using the RFID tag. The user of the IOT app can then review their bonus points, which can also be used to benefit from another discount (Chaturvedi et al., 2021). In the study "Development of Waste Management System Using the Concept of 'Basura Advantage Points' through Artificial Neural Network," researchers from Bulacan State University developed an automatic segregating device that employs Artificial Neural Network (ANN) as an algorithm for machine learning and is embedded with the concept of "Basura Advantage Points." The ANN acts as the machine's brain, categorizing waste materials according to their type. The "Basura Advantage Points" is a revolutionary idea that allows people to earn points for depositing trash in the garbage segregation machine. These points can subsequently be redeemed for rewards determined by lawmakers. The study brought to light a suitable model for reducing incorrect trash disposal and enticing people to engage in appropriate waste segregation (Castro et al., 2020).

## METHODOLOGY

### *Hardware and Software Development*

Figure 1 depicts the Agile Development approach that was adopted by the researchers to develop the device. It can help teams manage the work optimally while producing the best quality product within budget constraints. This method follows six fundamental phases, namely: requirements analysis, planning, design, coding, unit testing and acceptance from the client.

### *Plan*

In this phase the Gantt chart was developed to clearly specify the sequence of activities that were undertaken during the development of the study. It began with information gathering, which is the most vital activity that a project must perform to clearly identify the purpose and benefits of the project and ended with the submission of the hardbound manuscript. An outline of the project's requirements was included in the preliminary material that the proponents developed before work on the proposal really started. This phase entails reviewing and analyzing the requirements, such as the main variables of the study. By conducting a discussion, many ideas, and concepts on how the project's flow will work out are collected.

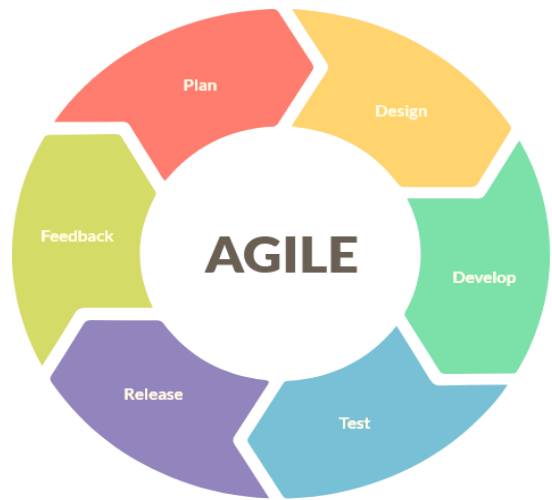

*Figure 1.* Agile Methodology

*Design*

In this phase, there were two approaches taken in designing the GULP smart bin: one was visual design, and the other was the architectural structure of the app in this phase, there were two approaches taken in designing the GULP smart bin: one was visual design, and the other was the architectural structure of the application. In the first iterative process, the project leader gathered the rest of the team and discussed the requirements developed during the requirement phase. The team then studied how to meet these objectives and proposed tools to achieve the best output. The System Design Architecture, Block Diagram, and Use Case Diagram were some of the tools used that guided the developers in realizing the architectural structure of the project. A rough mock-up of the user interface was also built by the project designer.

Figure 2 depicts the Block Diagram that begins with a solar panel added to the Smart Bin as its power source, so that the system is ready to operate. When in use, the user is only required to place garbage in the temporary storage box, after which a Pi camera controlled by the Raspberry Pi 4 will enable garbage classification. Each garbage container is labeled with a color: green for biodegradable, red for non-biodegradable, and yellow for recyclable. Each bin is represented with a light indicator that will perform in sync according to the type of garbage segregated. Furthermore, the stepper and servo motors will assist in transporting the garbage to its designated bin. The Dot Matrix LED module will display an emoticon after each use of the smart bin. Finally, the status of the bins will be determined by the ultrasonic sensors in each bin. The data will be sent back to the Raspberry Pi, where it will use the GSM/GPRS Module to generate an SMS alert message for the school's utility staff when it is time to clean the garbage.

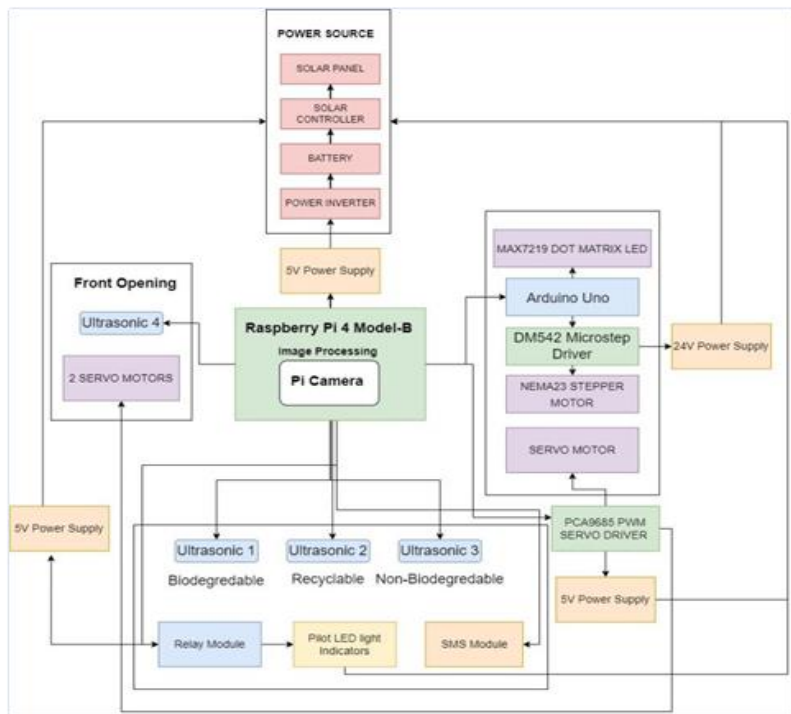

*Figure 2.* GULP Block Diagram

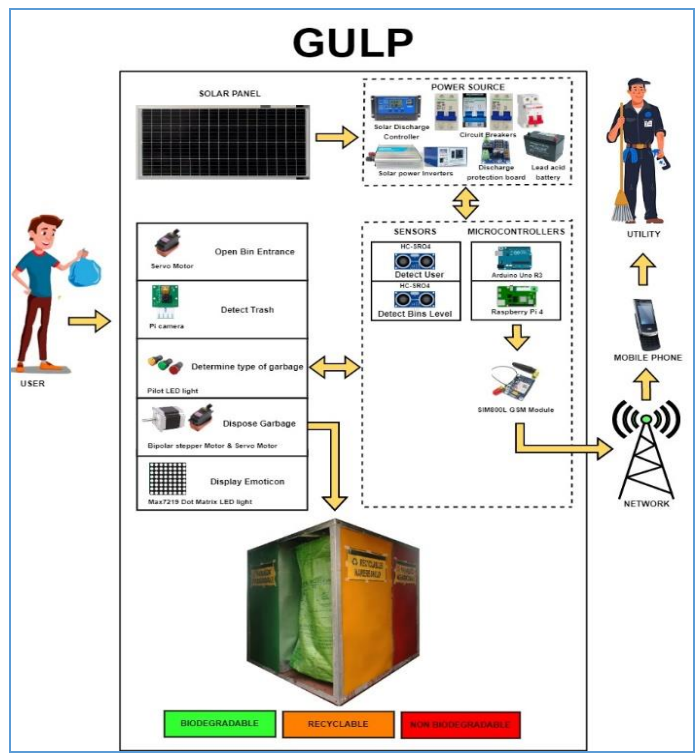

*Figure 3.* GULP System's Architecture

2026

As shown in Figure 3, the system's architecture organizes the main components needed to implement the application and the processes that occur during use. The physical machine is comprised of various hardware components integrated for a specific use. The hardware phase is divided into two sub-phases: 1) Front End (application interface the regular user would interact with, and 2) Back End (parts of the application and their code that allow it to operate and that cannot be accessed by the regular user like the above diagram with external components enclosed in broken lines). Once the user moves the garbage closes to the device, the bin lid will automatically open. Once the garbage is inside, the camera will take a photo of the trash that will be processed through the machine learning image processing. It will then be determined if it is biodegradable, non-biodegradable, and recyclable. The servo motor will rotate to the right trash bin, and the emoticon will display on the LED. Once the garbage reaches the full capacity, the system will automatically notify the utility staff through SMS. The device is solar powered and it also has a battery.

*Develop*

This phase includes the actual development that occurs after designing a prototype. In developing the application, researchers chose the Thonny Python IDE and Arduino IDE while using Python and Arduino programming language (framework based on the C++ language) as the main programming languages that allowed the researchers to inscribe computer commands. The development of the device required various hardware components such as the Raspberry Pi 4 Model B Starter Kit, which was used to run machine learning algorithms using its built-in camera input for image recognition. It accomplishes the task of recognizing objects and classifying them. Raspberry Pi 3 Model B+ Camera Module used to capture images and even high-definition videos for the machine learning. The system used a Convolutional Neural Network (ConvNet/CNN) for the image processing. It is a Deep Learning algorithm that can take in an input image, give importance (learnable weights and biases) to various aspects/objects in the image, and be able to distinguish one from the other. In comparison to other classification methods, a ConvNet requires significantly less pre-processing (Saha, 2018). Furthermore, Lobe.ai was used in training the data sets for the machine-learning models in image classification that utilized the Resnet50-V2 architectural building block with substantially higher accuracy. The 50-layer ResNet achieves a speed of 3.8 bn floating-point operations per second (FLOPS).

The Ultrasonic Sensor Distance Measuring Module HC-SR04 was applied to employ ultrasonic sound waves to gauge a distance to an item. It detects gesture to open the entrance door and for monitoring the bin level status. The SIM800L V2 5V Wireless GSM GPRS Module allows information to be transmitted via mobile networks. This is used as bearer of SMS. The Tower Pro Digital Robot Servo Motor (180 Rotation) – MG996R which is a self-contained electrical device rotates parts of a machine with high efficiency and with great precision. This is used both to open the entrance and to motion down the opening of the storage box used to temporarily store waste. Utilizing a modest light, the indicator signal LEDs are used to make the functioning state of the smart bins they are put on apparent from the outside.

To provide power for the smart bin, the system used a Solar Panel Mono-crystalline 200 Watt with MC4 Photovoltaic Connector. A 30A PWM 12V Solar Panel Regulator Charge Controller Solar Battery Charger LCD Display USB (COD) was employed to control the voltage and current flowing from the solar panel to the battery in order to prevent overcharging. The 105D31L Battery was used to store energy that will be used to power the smart bin at times when the solar panel will not generate enough electricity. The Solar panels produce direct current (DC) electricity, which was transformed into alternating current (AC) electricity using the 12V DC to 220-230V AC Car Home Solar Power Inverter with Buzzer. The Stepper Motor Nema 23 Wantai was used to deliver the waste to its proper bin. The MAX7219 Dot Matrix Module used to display symbols such as emoticon in the form of dots which consists of LED's. The 4 Channel 5V Relay Module with Optocoupler was used to switch devices which uses a higher voltage than what most micro-controllers such as a Raspberry Pi can handle. The power source that transforms AC power into the right kind of electricity for the device was the FiberSwitch Adaptor 5V 2A Heavy Duty. The circuit that monitors the battery voltage and disconnects the load to prevent deep discharge when the battery voltage falls below the lockout threshold is called the XH-M609 DC 12V 24V Voltage Charge Discharge Protection Board. Lastly, the UNO R3 Development Board ATmega328P CH340G was used to control stepper driver/motor and MAX7219 dot Matrix Module.

*Test*

This phase involves the product's development to test its functionality, and the researchers repeat the process until no errors are found. Table 1 shows the waste types tested which are few of the samples found in the dataset. The researchers conducted tests on various types of waste to determine whether the garbage was correctly sorted.

Figure 4 shows the data set training conducted by the researchers using Lobe.ai machine-learning image classification that utilized the Resnet50-V2 model. There are 1151 garbage images used with 98% of garbage images predicted correctly. The image processing was coded using Python programming language.

The performance of the application is assessed to see whether it complies with ISO/IEC 25010 Standards. Twenty-two (22) regular users, seven (7) utility personnel, and the Head of Health and Sanitation for a total of thirty (30) individuals evaluated the application.

The Likert scale that was utilized for the quantitative analysis is shown in Table 2. It is widely used to measure behavioral changes as well as attitudes, knowledge, beliefs, and values. An array of statements that respondents can select from to assess their answers to evaluative questions make up a Likert-type scale. The results and overall average of the ISO/IEC 25010 evaluation were revealed by the researchers using the Likert Scale.

Table 1. Types of Garbage Tested using GULP

| Garbage | Garbage Identification Result |
|---|---|
| Cardboard | Biodegradable |
| Yellow Paper | Biodegradable |
| Paper Meal Box | Biodegradable |
| Paper Bag | Biodegradable |
| Office Paper | Biodegradable |
| Newspaper | Biodegradable |
| Burger Paper Wrapper | Biodegradable |
| Milk tea Cup | Non-biodegradable |
| Instant Noodles Plastic Wrapper | Non-biodegradable |
| Cheese Snack Plastic Wrapper | Non-biodegradable |
| Ice Water Plastic Wrapper | Non-biodegradable |
| Juice Plastic Wrapper | Non-biodegradable |
| Coffee Stick Plastic Wrapper | Non-biodegradable |
| Chocolate Powder Plastic Wrapper | Non-biodegradable |
| Disposable Cups | Non-biodegradable |
| Candy Plastic Wrapper | Non-biodegradable |
| Corn Snack Plastic Wrapper | Non-biodegradable |
| Chocolate Bar Plastic Wrapper | Non-biodegradable |
| Biscuit Plastic Wrapper | Non-biodegradable |
| Black Plastic Trash Bag | Non-biodegradable |
| Instant Noodles Seasoning Plastic | Non-biodegradable |
| Juice Plastic Bottle | Recyclable |
| Water Plastic Bottle | Recyclable |
| Canned Goods Can | Recyclable |

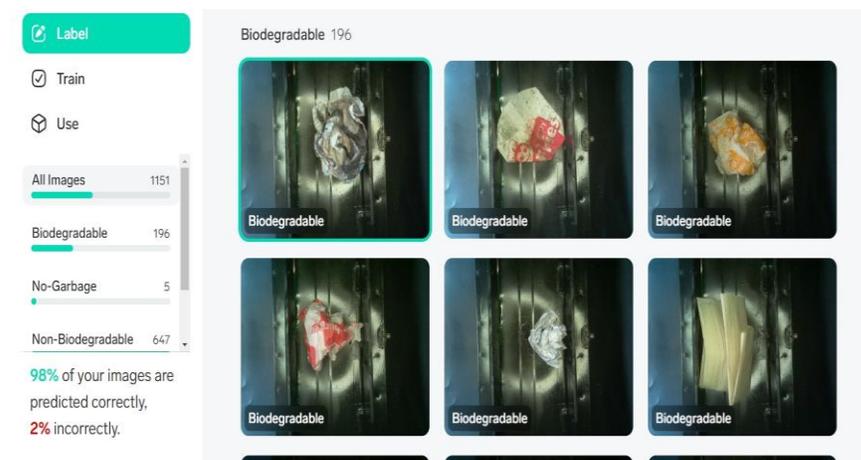

*Figure 4.* Garbage Data Set Training

Table 2. Likert Scale

| Scale | Parameters | Verbal Interpretation |
|---|---|---|
| 5 | 4.20 – 5.00 | Excellent (E) |
| 4 | 3.40 – 4.19 | Very Good (VG) |
| 3 | 2.60 – 3.39 | Good (G) |
| 2 | 1.80 – 2.59 | Fair (F) |
| 1 | 1.00 – 1.79 | Poor (P) |

*Release and Feedback*

The researchers presented the system to the organization for approval through presentation and system demonstration. A PowerPoint presentation was used for the preparatory and concise flow of the system while the actual system demonstration was done for genuine presentation.

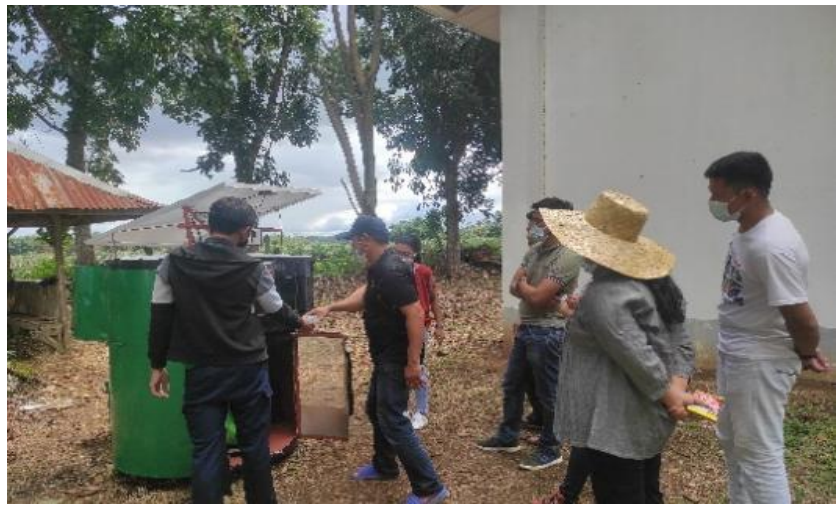

*Figure 5.* Training and Briefing

Before the training and briefing, the researchers set an appointment schedule on the presentation and demonstration where it includes the training and briefing of the application. During the presentation, the researchers conduct first the training and briefing to the users as shown in Figure 5. In the training and briefing part of the demonstration, the researchers had shown the application to the participants for them to understand how it works.

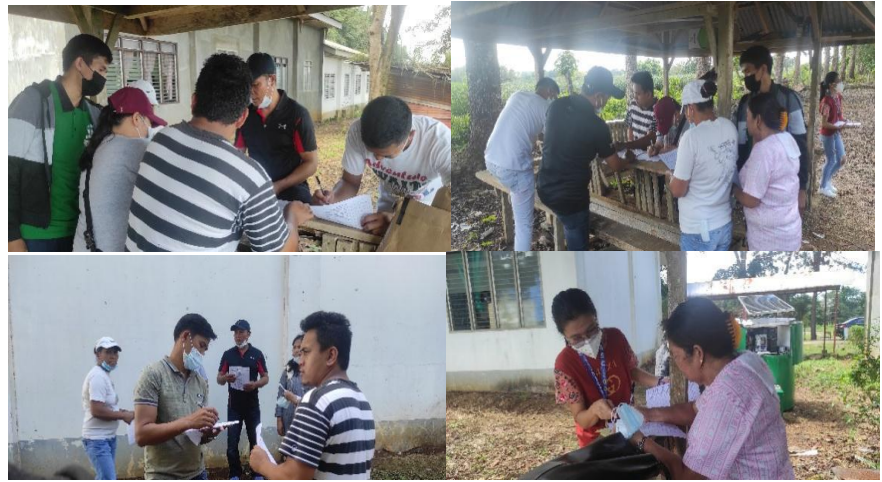

*Figure 6.* Evaluation and Feedback

Figure 6 shows the evaluation and feedback of the end-users with the system. They highly commend the functionality of the system, particularly on its mechanism to determine the type of garbage that will be thrown, notify the utility staff when it's full, and the displaying of emoticon. The utility staff were all amazed on how garbage disposal can be made fun and interactive for the users.

## RESULTS

Table 3 depicts the evaluation result that the researchers successfully designed the necessary requirements for the developed application through the use of Block Diagram that provided a functional view of the system, System Design Architecture for the operational mode on how it works, and Use-Case Diagram to depict user's possible interaction with the application. The developed application was evaluated and commended by the users using the ISO/IEC 25010. The functionality/suitability with a mean of 4.41, performance efficiency with a mean of 4.61, compatibility with a mean of 4.53, usability with a mean of 4.65, reliability with a mean of 4.52, security with a mean of 4.50, maintainability with a mean of 4.58, and portability with a mean of 4.57 which all resulted as "Excellent." The overall result Weighted Mean of 4.55 for functionality, performance efficiency, compatibility, usability, reliability, security, maintainability, and portability revealed an "Excellent" evaluation from the users.

Table 3. ISO/IEC 25010 Evaluation Result

| Characteristic | Mean | Verbal Interpretation |
|---|---|---|
| Functionality/ Suitability | 4.41 | Excellent (E) |
| Performance Efficiency | 4.61 | Excellent (E) |
| Compatibility | 4.53 | Excellent (E) |
| Usability | 4.65 | Excellent (E) |
| Reliability | 4.52 | Excellent (E) |
| Security | 4.50 | Excellent (E) |
| Maintainability | 4.58 | Excellent (E) |
| Portability | 4.57 | Excellent (E) |
| **Overall Weighted Mean** | **4.55** | **Excellent (E)** |

## DISCUSSION

The research gaps of the reviewed related studies helped the researchers to develop a smart and efficient waste disposal system that employs Convolutional Neural Network (ConvNet/CNN) deep learning algorithm using Lobe.ai machine-learning image classification that utilized the Resnet50-V2 model. This alleviates the developed system in faster and efficient garbage detection and segregation direct to an individual garbage bins that was suggested from a previous study. The machine learning image processing was coded using the python programming language, while the C++ programming language was utilized particularly on the Arduino interface of the microstep driver/stepper motor and the emoticon LCD display. The Short Message Service (SMS) that was employed to notify utility staff if the garbage bins are full addressed the research gap stated in one of the reviewed studies.

The excellent commendation of the end-users on its functionality and usability suffices the requirements that the developed system achieved the intended outcome. The system functions well in automatically opening the lid when a motion to throw garbage is detected. The system displays a countdown timer indicating that garbage recognition process is on-going. Once completed, the servo motor will direct intended garbage bin and the assigned light indicator LED will turn on. Then, the trash will be disposed. An emoticon will be displayed in the LCD. Which adds up to the fun for the users to enjoy the process of garbage disposal. The excellent commendation of the end-users for the performance efficiency, compatibility, reliability, and maintainability of the system, particularly on accurate detection of garbage classification for segregation will lighten up the workload of the utility workers in properly segregating the garbage since it will be directed to each separate bin as biodegradable, non-biodegradable, and recyclable, respectively.

## CONCLUSIONS AND RECOMMENDATIONS

Based on the result from the graph representation of the system, as rated by the respondents in terms of the functionality, suitability, performance efficiency, compatibility, usability, reliability, security, maintainability, and portability, the device is commended and accepted by the users of Northern Bukidnon State College. The gathered information is substantial for the researchers to conclude that GULP (Great User-oriented Litter Placer): Solar-Powered Smart Garbage Segregation Bins with SMS Notification and Machine Learning Image Processing is very significant, accessible, and convenient for use to dispose garbage and ensure proper waste disposal on campus. It is highly accepted with an excellent commendation by the end user using the ISO 25010 evaluation tool. Based on the findings and conclusions, the inclusion of compressors and a solar tracker coupled with solar panels is recommended for future researchers. A compressor will compress the trash when the trash level reaches its maximum point. Embedding it along with the system, will create more rooms for further disposal and save time and fuel, thus, waste collectors do not need to collect trash immediately. An algorithm to determine multiple garbage at a time is also recommended. Adding to the system, a solar tracker coupled with solar panel will help produce more renewable energy for the use of the smart bin. By rotating the solar panels to track the sun throughout the day and maximizing the angle at which they receive solar radiation, a solar tracking system will boost the solar array's ability to produce power.

## IMPLICATIONS

To prove that technological innovations and advancements can be utilized to make the waste segregation process more fun and interesting for the end-users. Currently, the institution continues to operate on the basis of traditional waste management along with the increasing student population, which entails an increasing effort in monitoring when waste is properly disposed. The traditional method of manually separating waste in a garbage bin is a time-consuming and costly process. Having effective and sustainable waste management application will help regulate waste disposal and alleviate some of the environmental pressures caused by consumption.

## ACKNOWLEDGEMENT

The researchers are truly indebted for all the commendable ideas, uplifting correction, and hands-on aid during the research phase. The researchers would like to give back all the glory and honor to God who has showered them with favor and sustained them all throughout and for providing them with people who have been instrumental to the success of this study.

# DECLARATIONS

## *Conflict of Interest*

The authors declare that there is no conflict of interest.

## *Informed Consent*

No direct private and personal information were used in the conduct of this research.

## *Ethics Approval*

As no private and personal information was used in the research, ethics approval is not necessary.

## Authors' Biography

Jerome B. Sigongan is a Bachelor of Science in Information Technology senior student at Northern Bukidnon State College, specializing in programming and internet of things.

Hamer P. Sinodlay is a Bachelor of Science in Information Technology senior student at Northern Bukidnon State College, specializing in hardware and internet of things.

Shahida Xerxy P. Cuizon is a Bachelor of Science in Information Technology senior student at Northern Bukidnon State College, specializing in human computer interaction and project management.

Joanna S. Redondo is a Bachelor of Science in Information Technology senior student at Northern Bukidnon State College, specializing in human computer interaction and project management.

Maricel G. Macapulay is a Bachelor of Science in Information Technology senior student at Northern Bukidnon State College, specializing in human computer interaction and project management.

Charlene O. Bulahan-Undag is an instructor at Northern Bukidnon State College, specializing in human computer interaction, professional ethics, internet of things and project management.

Kenn Migan Vincent C. Gumonan is the program head for the Bachelor of Science in Information Technology at Northern Bukidnon State College, specializing in research, programming, game development, internet of things and project management.